\documentclass[11pt,a4paper]{article}

\usepackage[margin=1in]{geometry}
\usepackage{graphicx}
\usepackage{booktabs}
\usepackage{hyperref}
\usepackage{amsmath}
\usepackage{amssymb}
\usepackage{enumitem}
\usepackage{xcolor}
\usepackage{xeCJK}
\usepackage{caption}

\setCJKsansfont{HaranoAjiGothic-Regular.otf}[BoldFont=HaranoAjiGothic-Bold.otf]
\setCJKmonofont{HaranoAjiMincho-Regular.otf}

\hypersetup{
    colorlinks=true,
    linkcolor=blue!50!black,
    citecolor=blue!50!black,
    urlcolor=blue!50!black,
}

\title{Retrieval and Multi-Hop Reasoning in 1M-Token Context Windows:\\
Evaluating LLMs on Classical Chinese Text}

\author{Eric H. C. Chow\\
School of Humanities, The University of Hong Kong\\
\texttt{eric.chow@hku.hk}}
\date{May 2026}

\begin{document}

\maketitle

\begin{abstract}
We evaluate the long-context retrieval and reasoning capabilities of five frontier large language models with advertised 1M-token context windows on a classical Chinese (文言文) corpus. Two complementary studies are reported. \textbf{Test 1} measures single-needle retrieval at 1M tokens of input, with three biographical needles planted at three depths and pairs of \emph{real} (training-prior-consistent) and \emph{altered} (training-prior-contradicting) variants to separate genuine in-context retrieval from reliance on memorised training data. \textbf{Test 2}, a follow-up designed to probe whether long-context capability degrades when retrieval requires intermediate reasoning, measures three-hop chain traversal across three context tiers (256K, 512K, and 1M tokens). We find that single-needle retrieval at 1M is essentially solved for the strongest models~--- Gemini 3.1 Pro, Claude Opus 4.7, and GPT-5.5 each achieve 100\%~--- but that multi-hop performance reveals three distinct decay signatures: a \emph{stable} regime (Gemini Pro, Claude) maintaining $\geq$80\% accuracy through 512K with modest degradation at 1M; a \emph{late-cliff} regime (GPT-5.5, Qwen3.6-plus) collapsing sharply between 512K and 1M (4/5 $\to$ 2/5 and 4/5 $\to$ 0/5 respectively); and a \emph{smooth-decline} regime (DeepSeek V4 Pro) decaying gradually across the entire range. The findings suggest that nominal context-window length is a poor proxy for usable long-context multi-hop capability, and that the sharpest discriminator between current 1M-context flagships is the 512K-to-1M transition.

\end{abstract}

\section{Introduction}

Frontier large language models (LLMs) have rapidly expanded their advertised context windows from 4K tokens in 2022 to 1M+ tokens in 2026. This expansion raises a practical question for anyone building question-answering systems over document corpora: \emph{does a sufficiently large context window make RAG unnecessary?} If a model can ingest an entire document corpus in a single prompt and answer questions faithfully, the retrieval stage~--- whether traditional vector-similarity RAG or the more recent Graph RAG approach that first extracts a knowledge graph from the corpus~--- becomes redundant overhead. Conversely, if long-context attention degrades on the kinds of multi-hop, relation-traversal queries that motivate Graph RAG in the first place, then a large context window alone is not enough, and a separate retrieval step still earns its keep.

This question is especially relevant for historians working with large pre-modern text corpora. A scholar studying Song-dynasty intellectual networks, for instance, might want to query the full text of the \emph{Song-Yuan Xue'an} (《宋元學案》)~--- a 1.4-million-character compilation of Neo-Confucian biographical and scholarly records~--- for prosopographical details: who studied under whom, which scholars were contemporaries, what offices a minor figure held. These are precisely the kinds of single-fact lookups and multi-hop relationship queries that RAG systems are designed to support. If a 1M-context model can answer them directly from the raw text, it simplifies the historian's workflow considerably; if it cannot, the historian needs a structured retrieval layer~--- and knowing \emph{where} the model breaks helps determine what kind.

We approach this question empirically. The standard \emph{needle-in-a-haystack} (NIAH) benchmark measures whether a model can locate a planted statement within a long document; it has been widely applied to English text and modern Chinese, but classical Chinese (文言文) presents a distinct evaluation regime for three reasons:

\begin{enumerate}[label=\arabic*.]
\item \textbf{Tokenization asymmetry.} Modern subword tokenizers, optimized largely on English and modern languages, fragment classical-Chinese characters at very different rates across vendors. The same character corpus may consume 0.9--1.2 tokens per character depending on the model, with substantial implications for usable context length.
\item \textbf{Parametric leakage.} Many classical-Chinese texts appear in LLM pretraining corpora, either directly or through secondary scholarship. A model may therefore appear to ``retrieve'' a fact from the provided context when it is actually drawing on memorised training data, masking failures in long-context attention.
\item \textbf{Compactness and ambiguity.} Classical-Chinese characters carry dense semantic content per character, magnifying both the effective information rate of the haystack and the cost of mis-attention to a single token.
\end{enumerate}

We design two complementary studies that together probe (a) whether the strongest current 1M-context models can locate and report facts planted in the provided text~--- even when those facts contradict what the model has seen in training~--- and (b) whether this retrieval capability degrades smoothly with context length or fails sharply at a particular threshold.

The contributions of this paper are:
\begin{itemize}[leftmargin=*]
\item A controlled multi-model evaluation on a classical-Chinese corpus designed to separate genuine in-context retrieval from reliance on memorised training data, via paired \emph{real}/\emph{altered} needle variants;
\item A multi-hop chain task using fictional subject names, eliminating training-data leakage entirely and forcing genuine in-haystack reference traversal;
\item A three-tier (256K / 512K / 1M) accuracy curve for each model, revealing three qualitatively distinct decay signatures across five frontier flagships.
\end{itemize}

\section{Related Work}

\subsection{Long-Context Retrieval Benchmarks}

The needle-in-a-haystack (NIAH) test was introduced by Kamradt (2023) as an informal stress test: plant a single statement in a long document and check whether the model can retrieve it. While effective as a basic verification tool, subsequent work has exposed its limitations. RULER (Hsieh et al., 2024), published at COLM 2024, extended the vanilla NIAH paradigm with four task categories~--- retrieval, multi-hop tracing, aggregation, and question answering~--- and found that despite achieving near-perfect NIAH accuracy, ``almost all models exhibit large performance drops as the context length increases.'' Critically, while the 17 models tested all claimed context sizes of 32K or greater, only half maintained satisfactory performance at that length. Beyond retrieval, RULER also introduced a multi-hop tracing task that tracks variable assignments across the context, though this operates on synthetic English-language variable-binding.

BABILong (Kuratov et al., 2024), presented at NeurIPS 2024, scaled the evaluation further to multi-fact reasoning tasks. Its key finding~--- that ``popular LLMs effectively utilize only 10--20\% of the context and their performance declines sharply with increased reasoning complexity''~--- suggests that the gap between advertised and functional context length is wide. BABILong also found that Retrieval-Augmented Generation methods achieve only modest 60\% accuracy on single-fact questions, independent of context length. However, BABILong uses synthetic English facts embedded in fiction prose (PG19 novels), leaving open the question of whether these findings generalise to real-world corpora with domain-specific semantics and register.

NeedleBench (Li et al., 2024) added bilingual (Chinese--English) evaluation and introduced an ``Ancestral Trace Challenge'' requiring multi-step reasoning over continuously distributed biographical information. Sequential-NIAH (Yu et al., 2025) tested extraction of multiple ordered items from 8K--128K contexts and found that even the best model achieved only 63.5\% accuracy, confirming that multi-element retrieval remains substantially harder than single-needle lookup.

Several gaps in this literature motivate our study. First, no existing NIAH-family benchmark tests at 1M tokens on the current generation of frontier models (April 2026). RULER stops at 128K; BABILong evaluated older models. Second, all existing benchmarks use English or modern Chinese~--- none uses classical Chinese, a domain with distinct tokenization characteristics, high information density, and substantial parametric-leakage risk. Third, existing multi-hop tasks either use synthetic variable-tracking (RULER) or fictional English-language facts (BABILong), neither of which captures the relational, graph-traversal structure of the entity-and-relation queries that motivate GraphRAG-style systems~--- the kind of query a historian might ask of a biographical corpus, but equally relevant to legal, biomedical, or enterprise document collections. The present study addresses all three gaps.

\subsection{RAG Versus Long Context}

The question of whether expanding context windows renders RAG unnecessary has been addressed directly in recent work. Li et al.\ (2025a) compared long-context and RAG approaches on question-answering benchmarks and found that long context generally outperforms RAG for Wikipedia-based queries, though RAG retains advantages in dialogue settings and offers better cost efficiency. The LaRA benchmark (Li et al., 2025b), presented at ICML 2025, evaluated 11 LLMs across 2,326 test cases and concluded that ``neither RAG nor long-context LLMs are a silver bullet''~--- the optimal choice depends on model capability, context length, task type, and retrieval quality. Neither study, however, characterises the \emph{shape} of long-context decay across tiers or offers guidance on \emph{which} models benefit from RAG at \emph{which} context lengths~--- a gap that a multi-tier evaluation design could fill.

For multi-hop queries specifically, Edge et al.\ (2024) proposed GraphRAG, which first extracts an entity knowledge graph from the corpus and then traverses it to answer relational queries. GraphRAG demonstrated substantial improvements over baseline RAG for global sensemaking questions on million-token corpora. Multi-hop chain queries~--- of the kind common in prosopographical research~--- are structurally equivalent to GraphRAG traversal queries, raising the question of when such a preprocessing stage adds value versus when long-context attention alone suffices.

\subsection{Classical Chinese NLP}

Classical Chinese has emerged as a distinct evaluation target for LLMs, motivated by the recurring observation that models pre-trained primarily on modern Chinese transfer poorly to the older, written-only style of Chinese used in pre-modern texts. Three benchmarks anchor this literature. WYWEB (Zhou et al., 2023), published at ACL 2023 Findings, assembles nine sentence- and paragraph-level tasks~--- punctuation restoration, named entity recognition, sentence classification by genre and historical period, sentiment recognition for ancient poetry, multiple-choice reading comprehension, function-word disambiguation, and machine translation into modern Chinese~--- and reports that existing pre-trained models ``all struggled,'' attributing the gap to a fundamental mismatch between the modern Chinese the models were trained on and the classical Chinese the benchmark tests. WenMind (Cao et al., 2024, NeurIPS Datasets \& Benchmarks) extends the evaluation surface to 4{,}875 question--answer pairs across 42 tasks spanning ancient prose, ancient poetry, and ancient literary culture, grouped along three capability axes (understanding, generation, knowledge). Its headline result is that even the strongest model evaluated (ERNIE-4.0) reaches only 64.3 overall while most models score between 20 and 60, and that knowledge-focused tasks~--- particularly biographical and literary-culture questions~--- exhibit the largest deficits relative to understanding and generation. WenyanGPT (Yao et al., 2025) responds to these gaps by continually pre-training a LLaMA3-8B-Chinese model on a 16~GB classical-Chinese corpus and instruction-tuning it on six tasks (punctuation, POS tagging, NER, translation, word explanation, reverse dictionary), demonstrating that targeted adaptation can close much of the gap between modern and classical Chinese on sentence-level diagnostics.

Three observations from this literature directly inform our design. First, all three benchmarks operate at sentence- or paragraph-length inputs, with WenMind question--answer pairs averaging under 80 tokens and WYWEB samples drawn at the paragraph level; none probes whether the same models can locate information embedded in book-length classical-Chinese contexts. Second, WenMind's documented \emph{knowledge deficit} on biographical and literary-culture questions is measured against a model's parametric memory~--- whether providing the source text in-context closes that gap, and whether models can compose relations between figures rather than recall isolated facts, has not been tested. Third, WenMind's fine-grained tasks on polysemy, function words, and altered sentence structures confirm that small-character substitutions in classical Chinese are linguistically meaningful and already difficult at the sentence level; our altered-needle variants extend that probe to a one-million-token scale, where context dilution and the tokenization asymmetry described in §2.1 compound the difficulty. Long-context retrieval and multi-hop relational reasoning on classical Chinese~--- the kind of query a historian, lexicographer, or cultural-heritage retrieval system would naturally pose over an indexed corpus~--- remain unaddressed by existing benchmarks, and the two tests reported below are designed to fill exactly this slot.

\section{Methods}

\subsection{Corpus}

Both tests use the \emph{Song-Yuan Xue'an} (《宋元學案》), a Qing-dynasty compilation of biographical and intellectual records of Song- and Yuan-era Confucian scholars. We use the digital edition hosted by the \emph{Chinese Text Project} (\url{https://ctext.org/wiki.pl?if=gb&res=370460}). The compilation comprises approximately 1.42 million characters across 100 volumes (卷), of which we use the 97 volumes that are not the personal 學案 of any needle subject.

Volumes are concatenated in canonical 卷 order to form a \emph{master haystack}. The full master is then sliced into per-tier haystacks (described in §3.4).

\subsection{Models}

We evaluate five LLMs with 1M-token context windows. All five are frontier models at the time of writing, released between October 2025 and April 2026:

\begin{table}[h]
\centering
\small
\begin{tabular}{@{}llr@{}}
\toprule
\textbf{Model} & \textbf{Provider} & \textbf{Tokenizer} \\
\midrule
Gemini 3.1 Pro     & Google     & SentencePiece    \\
Claude Opus 4.7    & Anthropic  & BPE              \\
GPT-5.5            & OpenAI     & o200k\_base BPE  \\
Qwen3.6-plus       & Alibaba    & Qwen3 BPE        \\
DeepSeek V4 Pro    & DeepSeek   & DeepSeek BPE     \\
\bottomrule
\end{tabular}
\caption{Models evaluated. All five advertise context windows of approximately 1M tokens.}
\end{table}

All five are accessed via official APIs. Where the API allows, sampling temperature is set to 0 to maximize determinism; Claude Opus 4.7 and GPT-5.5 enforce temperature\,=\,1 when reasoning is enabled, and we report results at this constraint. Reasoning effort or thinking mode is set to the model's highest available level for thinking-capable models, to ensure each model receives an equal computational budget.

\subsection{Tokenization Calibration}

Each model's tokenizer was calibrated against the corpus to determine its tokens-per-character ratio:

\begin{table}[h]
\centering
\small
\begin{tabular}{@{}lr@{}}
\toprule
\textbf{Model} & \textbf{Tokens / character} \\
\midrule
DeepSeek V4 Pro    & 0.901 \\
Qwen3.6-plus       & 0.917 \\
Gemini 3.1 Pro     & 0.945 \\
GPT-5.5            & 1.040 \\
Claude Opus 4.7    & 1.222 \\
\bottomrule
\end{tabular}
\caption{Per-model tokenization rate on the classical Chinese corpus.}
\end{table}

The 35\% gap between the most and least efficient tokenizers implies that a haystack sized to fill one model's context window will substantially under- or over-fill another's. We therefore construct \textbf{per-model haystacks}, sized via each model's own tokenizer to consume approximately 99\% of its usable input ceiling. For Test 2's lower tiers, an analogous procedure is applied at 256K- and 512K-token targets.

\subsection{Test 1: Single-Needle Retrieval at 1M Tokens}

\begin{figure}[h]
\centering
\includegraphics[width=0.95\textwidth]{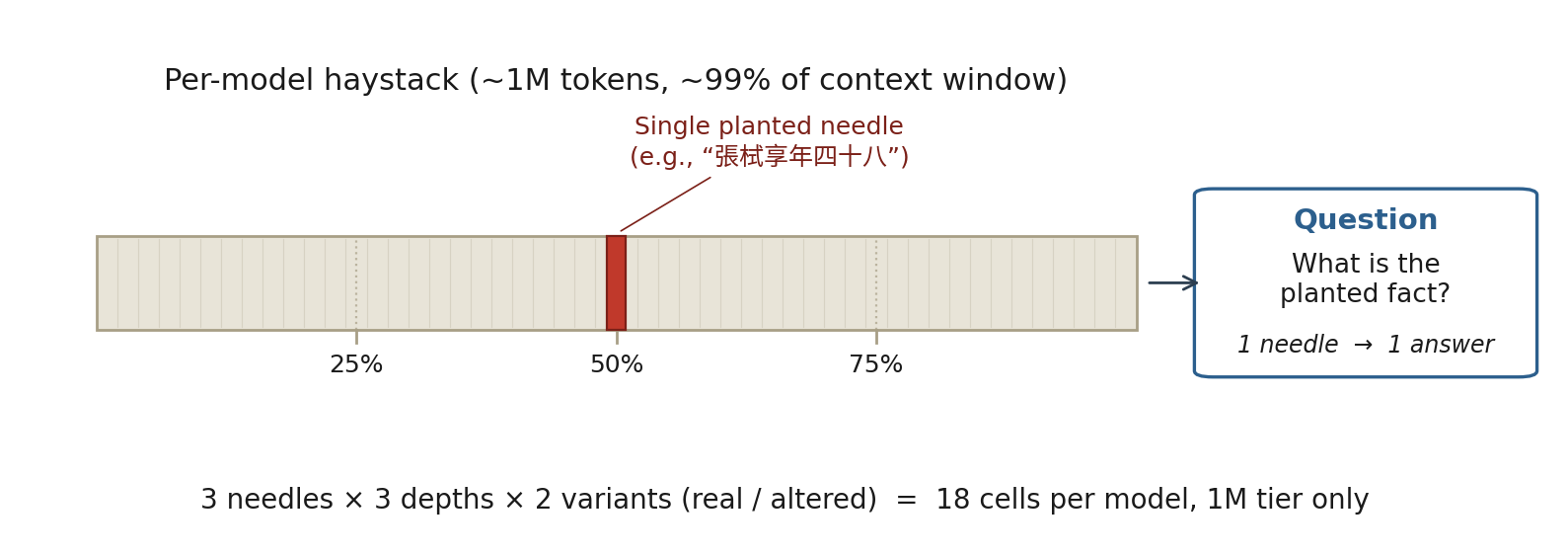}
\caption{Test 1 schematic. A single biographical fact is planted at one of three depths (25\%, 50\%, 75\%) inside a per-model haystack sized to $\sim$99\% of each model's 1M-token context window. The model is then asked one question whose answer depends solely on the planted statement.}
\label{fig:test1-diagram}
\end{figure}

Three biographical needles were constructed, each based on a verifiable historical fact about a Southern Song scholar whose own biographical 學案 is excluded from the haystack:

\begin{table}[h]
\centering
\small
\begin{tabular}{@{}llllll@{}}
\toprule
\textbf{ID} & \textbf{Subject} & \textbf{Question} & \textbf{Real} & \textbf{Altered} \\
\midrule
N1 & 張栻 (Zhang Shi)     & Age of death          & 48 (四十八)     & 53 (五十三)     \\
N2 & 真德秀 (Zhen Dexiu)  & Place of origin    & 浦城            & 建陽            \\
N3 & 黃榦 (Huang Gan)     & Kinship to Zhu Xi  & 婿              & 甥              \\
\bottomrule
\end{tabular}
\caption{Test 1 needle definitions.}
\end{table}

The \emph{altered} values are deliberately constructed to contradict what the model would have encountered in training. For each (needle $\times$ variant), the planted statement is inserted at one of three depths~--- 25\%, 50\%, or 75\% of paragraph index~--- yielding $3 \times 3 \times 2 = \textbf{18 cells per model}$, run at the 1M tier exclusively.

Cross-references to the needle subjects in the haystack are unavoidable but the specific planted facts (death age, place of origin, kinship) appear nowhere outside the planted needle. This was verified by exhaustive search of the master haystack against each fact.

\subsection{Test 2: Multi-Hop Chain Retrieval Across Three Tiers}

\begin{figure}[h]
\centering
\includegraphics[width=0.95\textwidth]{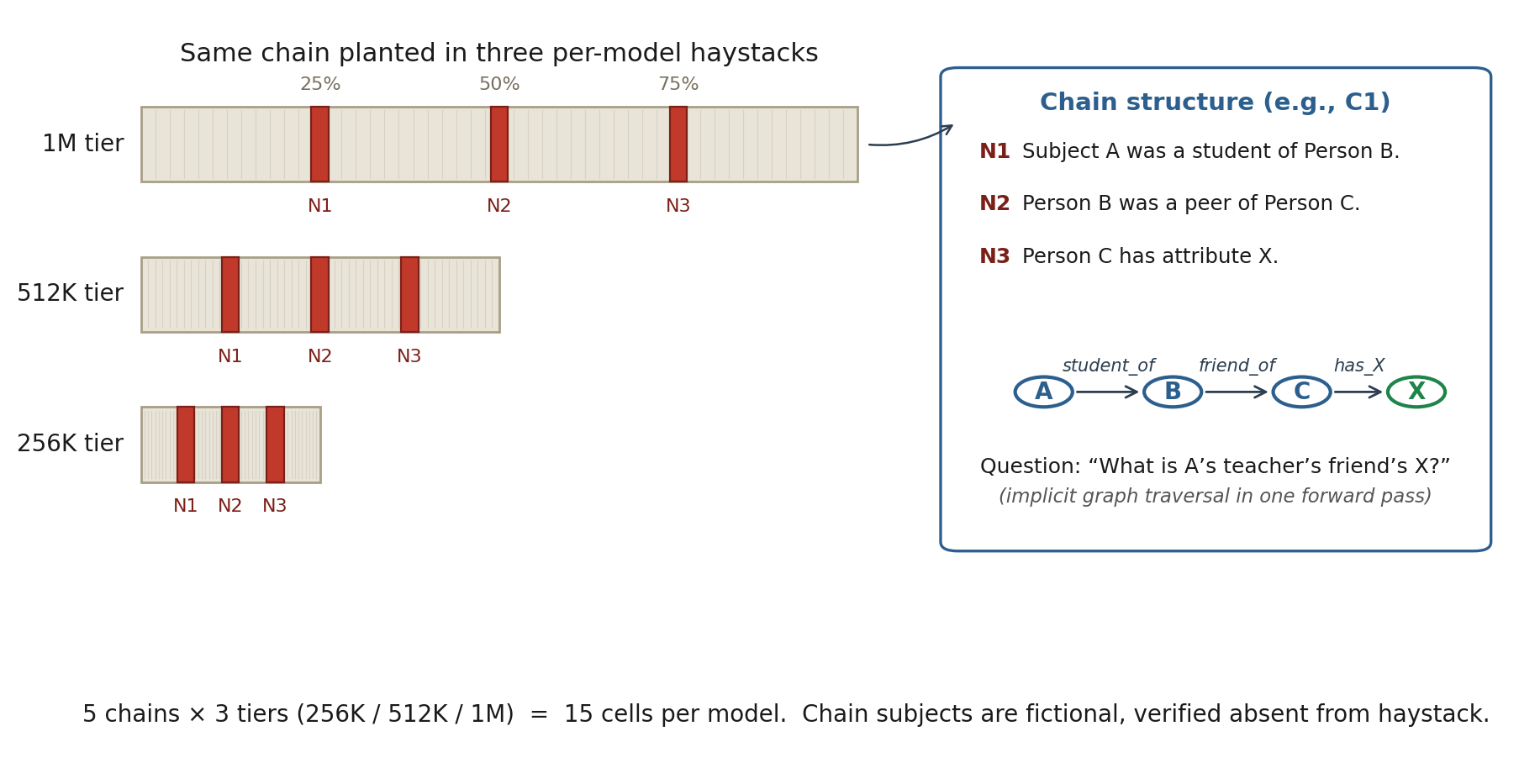}
\caption{Test 2 schematic. Each chain consists of three planted statements (N1, N2, N3) inserted at fixed depths of 25\%, 50\%, and 75\% in a per-model haystack at three tiers (256K / 512K / 1M). The single question requires the model to traverse the relations $A \to B \to C$ and then read $C$'s attribute $X$, an implicit graph traversal performed in one forward pass.}
\label{fig:test2-diagram}
\end{figure}

Test 2 strengthens the test along two dimensions: it requires \emph{traversing references} rather than locating a single fact, and it uses \emph{fictional subject names} to fully eliminate training-data leakage. Where prior multi-hop benchmarks use synthetic variable-tracking (RULER) or English-language fact chaining (BABILong), our chains are modelled on the relational structure of prosopographical queries~--- the kind of ``who studied under whom'' questions that historians actually ask of biographical corpora.

Five three-hop chains were constructed, each consisting of three planted statements that together form a referential chain:

\begin{quote}
\textit{N1: Subject A was a student of Person B.}\\
\textit{N2: Person B was a friend (or peer) of Person C.}\\
\textit{N3: Person C has the attribute X.}\\
Question: \textit{What is Subject A's teacher's friend's X?}
\end{quote}

The five chains test five different attribute types. All subject and intermediate names are fictional, with verified absence from the master haystack. The chains are summarised in Table~\ref{tab:chains}.

\begin{table}[h]
\centering
\small
\begin{tabular}{@{}clllllll@{}}
\toprule
\textbf{Chain} & \textbf{A (start)} & \textbf{A$\to$B} & \textbf{B} & \textbf{B$\to$C} & \textbf{C} & \textbf{Attribute of C} & \textbf{Expected} \\
\midrule
C1 & 陳遵叔 & student of & 劉允之 & classmate of & 羅彥脩 & age at death & 61 \\
C2 & 孫有開 & student of & 張紹卿 & friend of    & 葉與時 & year of death & 紹熙四年 \\
C3 & 蔡守端 & student of & 何叔諒 & friend of    & 韓彥行 & hometown & 吳縣 \\
C4 & 鄭子潛 & uncle      & 陳叔卿 & friend of    & 張延猷 & book authored & 中庸辨疑 \\
C5 & 黃師復 & student of & 馬德明 & co-local of  & 李允升 & courtesy name & 明甫 \\
\bottomrule
\end{tabular}
\caption{The five three-hop chains used in Test 2. All person names (A, B, C) are fictional. The question asks for C's attribute given only A's name.}
\label{tab:chains}
\end{table}

For each chain, the three needles are inserted into the haystack at fixed depths of 25\%, 50\%, and 75\% of paragraph index. The model is then asked the chain question. With $5 \times 3 = \textbf{15 cells per model}$, totaling 75 cells across the five models.

The three tiers are 256K (control), 512K (midpoint), and 1M (target). Per-model haystacks are constructed for each tier such that each haystack consumes approximately 99\% of the corresponding token budget on the model's own tokenizer.

\subsubsection{Chain Task as Implicit Knowledge-Graph Traversal}
\label{sec:kg}

Each chain is, structurally, a minimal knowledge graph (KG). The three planted statements correspond to three RDF-style triples:

\begin{verbatim}
(Subject A, student_of, Person B)
(Person B,  friend_of,  Person C)
(Person C,  has_X,      attribute_value)
\end{verbatim}

The question --- ``What is \emph{Subject A}'s teacher's friend's \emph{X}?'' --- is a graph traversal query: starting at node \emph{A}, follow the \texttt{student\_of} edge to \emph{B}, then the \texttt{friend\_of} edge to \emph{C}, then read the \texttt{has\_X} attribute. In a Graph Retrieval-Augmented Generation (GraphRAG) system, this query would be answered in two explicit stages: (i) an extraction stage that parses the document corpus into a typed graph of entities and relations, and (ii) a traversal stage that walks the graph along the relation path implied by the question.

In Test 2, the language model performs both stages \emph{implicitly}~--- it must (a) recognise that the three planted statements form a chain, (b) bind the intermediate entity names across statements, and (c) extract the final attribute, all within a single forward pass over the haystack. Test 2 therefore measures the upper bound of what a long-context LLM can do \emph{without} an external KG: the score is the proportion of multi-hop graph traversal queries that long-context attention can resolve end-to-end.

The three-hop depth of the chains is the minimum non-trivial graph traversal --- single-hop reduces to Test 1's single-needle retrieval, while two-hop chains lack an explicit \emph{intermediate-binding} test. Three hops force the model to maintain at least one intermediate referent (Person B and Person C) in working memory while continuing to scan the haystack for the next link in the chain.

\subsection{Scoring}

For both tests, a response is scored as correct if the expected value appears as a substring of the model's reply. Numeric answers are accepted in either Arabic or Chinese-numeral form (e.g., ``61'' matches ``六十一''). The model is free to answer in any form~--- no format constraints are imposed.

\subsection{Implementation}

All models are queried via their official API endpoints. Responses include both visible answer content and reasoning content (where the API exposes the latter as plaintext); both are recorded for each cell.

\section{Results}

\subsection{Test 1: Single-Needle Retrieval}

\begin{table}[h]
\centering
\small
\begin{tabular}{@{}lrr@{}}
\toprule
\textbf{Model} & \textbf{Score} & \textbf{Accuracy} \\
\midrule
Gemini 3.1 Pro     & 18 / 18 & 100\% \\
Claude Opus 4.7    & 18 / 18 & 100\% \\
GPT-5.5            & 18 / 18 & 100\% \\
DeepSeek V4 Pro    & 13 / 18 & 72\%  \\
Qwen3.6-plus       &  7 / 18 & 39\%  \\
\bottomrule
\end{tabular}
\caption{Test 1 results: 5 models $\times$ 18 cells (3 needles $\times$ 3 depths $\times$ 2 variants) at 1M tokens.}
\end{table}

\begin{figure}[h]
\centering
\includegraphics[width=0.85\textwidth]{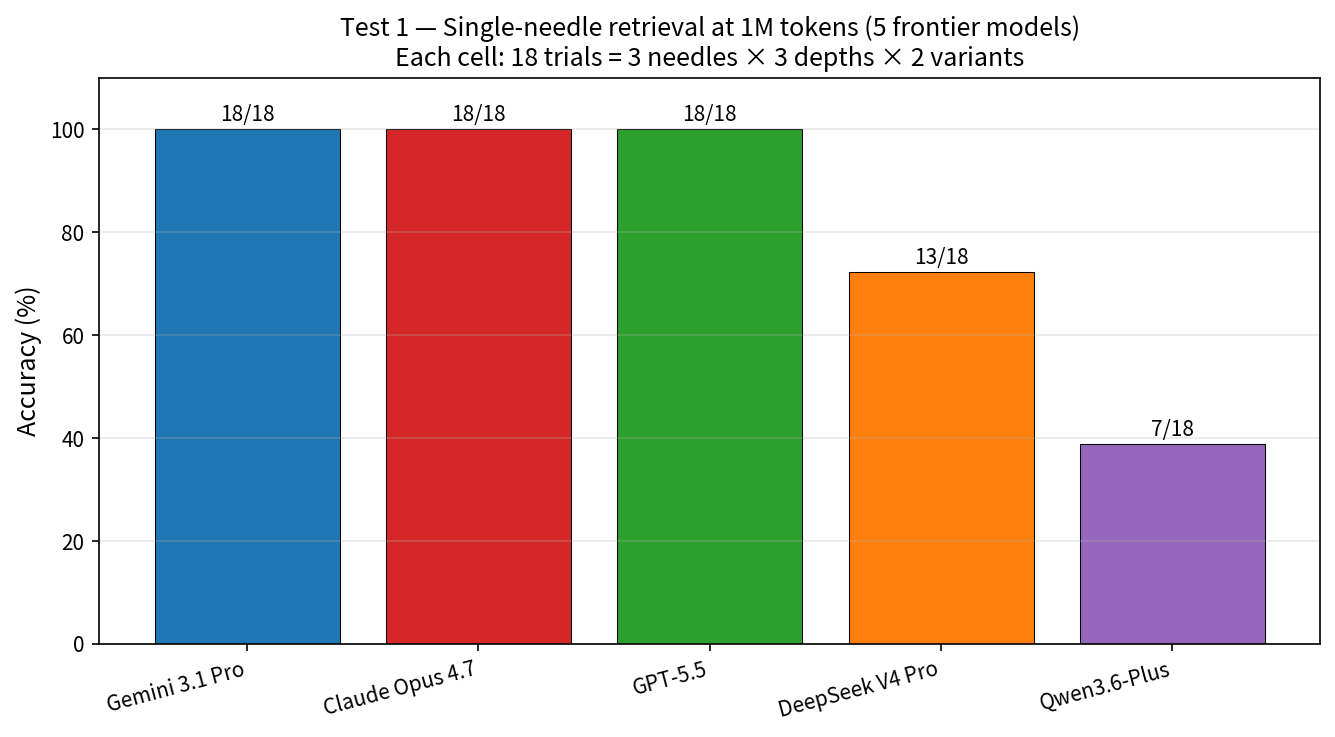}
\caption{Single-needle retrieval accuracy at 1M tokens, ordered by score.}
\end{figure}

As shown in Figure~1, the three top-tier models~--- Gemini 3.1 Pro, Claude Opus 4.7, and GPT-5.5~--- achieve perfect retrieval across all 18 cells, including all altered-variant cells where the planted statement contradicts what the model would have seen in training. This indicates that, at single-fact granularity, these models successfully attend to and report content from the provided text rather than defaulting to memorised training data.

DeepSeek V4 Pro achieves 13/18; failure analysis (Figure~2) shows that its errors cluster on the \emph{altered} variants of N2 (place of origin), where the model returns the training-consistent answer (浦城) at all three depths despite the haystack containing the contradicting needle. This is consistent with the model falling back to memorised biographical data about Zhen Dexiu rather than reading the provided text.

Qwen3.6-plus achieves 7/18, but inspection of its reasoning content reveals that the majority of its successes are accidental: the model's reply pattern is overwhelmingly ``the haystack does not record this fact, but per historical record the answer is $X$,'' where $X$ happens to coincide with the expected value on real-variant cells but never with the altered-variant value. Across all six altered-variant cells, Qwen3.6-plus achieves 0/6, supporting the interpretation that its 1M-context retrieval is dominated by memorised training data rather than in-context attention.

\begin{figure}[h]
\centering
\includegraphics[width=\textwidth]{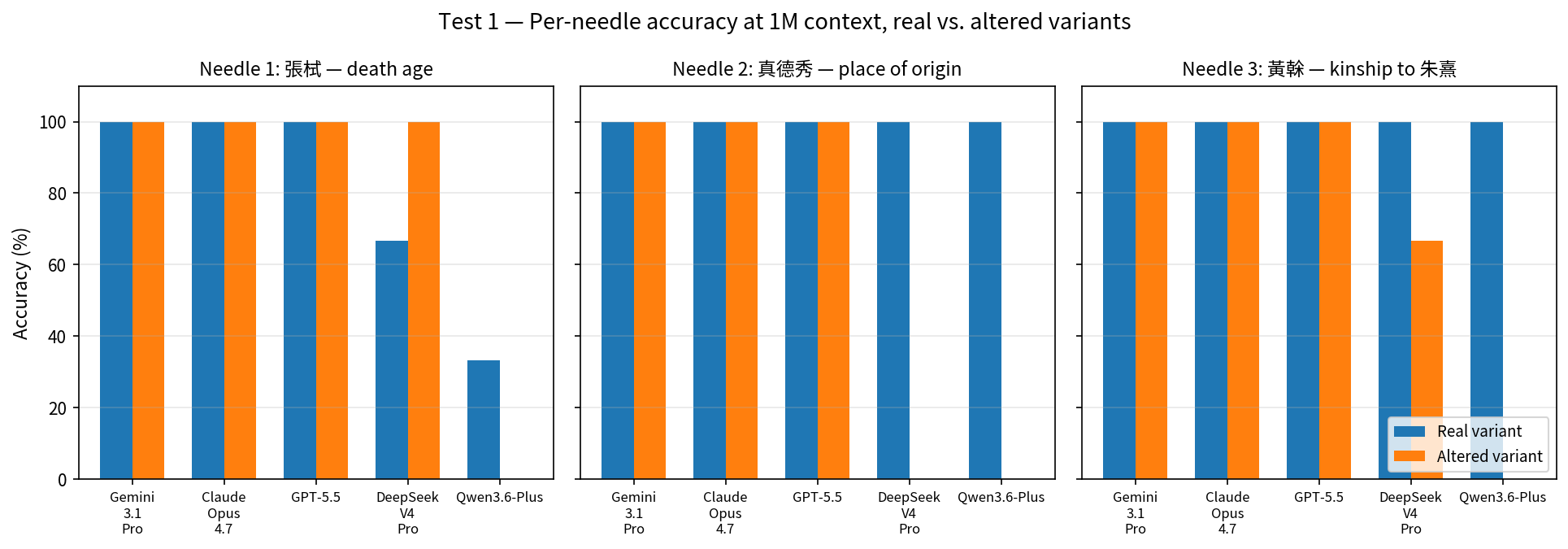}
\caption{Per-needle accuracy split by variant. Altered-variant accuracy provides the cleaner signal of genuine in-context retrieval.}
\end{figure}

\subsection{Test 2: Multi-Hop Chain Retrieval}

\begin{table}[h]
\centering
\small
\begin{tabular}{@{}lrrrr@{}}
\toprule
\textbf{Model} & \textbf{256K} & \textbf{512K} & \textbf{1M} & \textbf{Total} \\
\midrule
Gemini 3.1 Pro     & 5/5 & 5/5 & 5/5  & 15/15 (100\%) \\
Claude Opus 4.7    & 4/5 & 5/5 & 3/5  & 12/15 (80\%) \\
GPT-5.5            & 5/5 & 4/5 & 2/5  & 11/15 (73\%) \\
Qwen3.6-plus       & 5/5 & 4/5 & 0/5  &  9/15 (60\%) \\
DeepSeek V4 Pro    & 4/5 & 2/5 & 0/5  &  6/15 (40\%) \\
\bottomrule
\end{tabular}
\caption{Test 2 results: 5 models $\times$ 15 cells (5 chains $\times$ 3 tiers).}
\end{table}

\begin{figure}[h]
\centering
\includegraphics[width=0.95\textwidth]{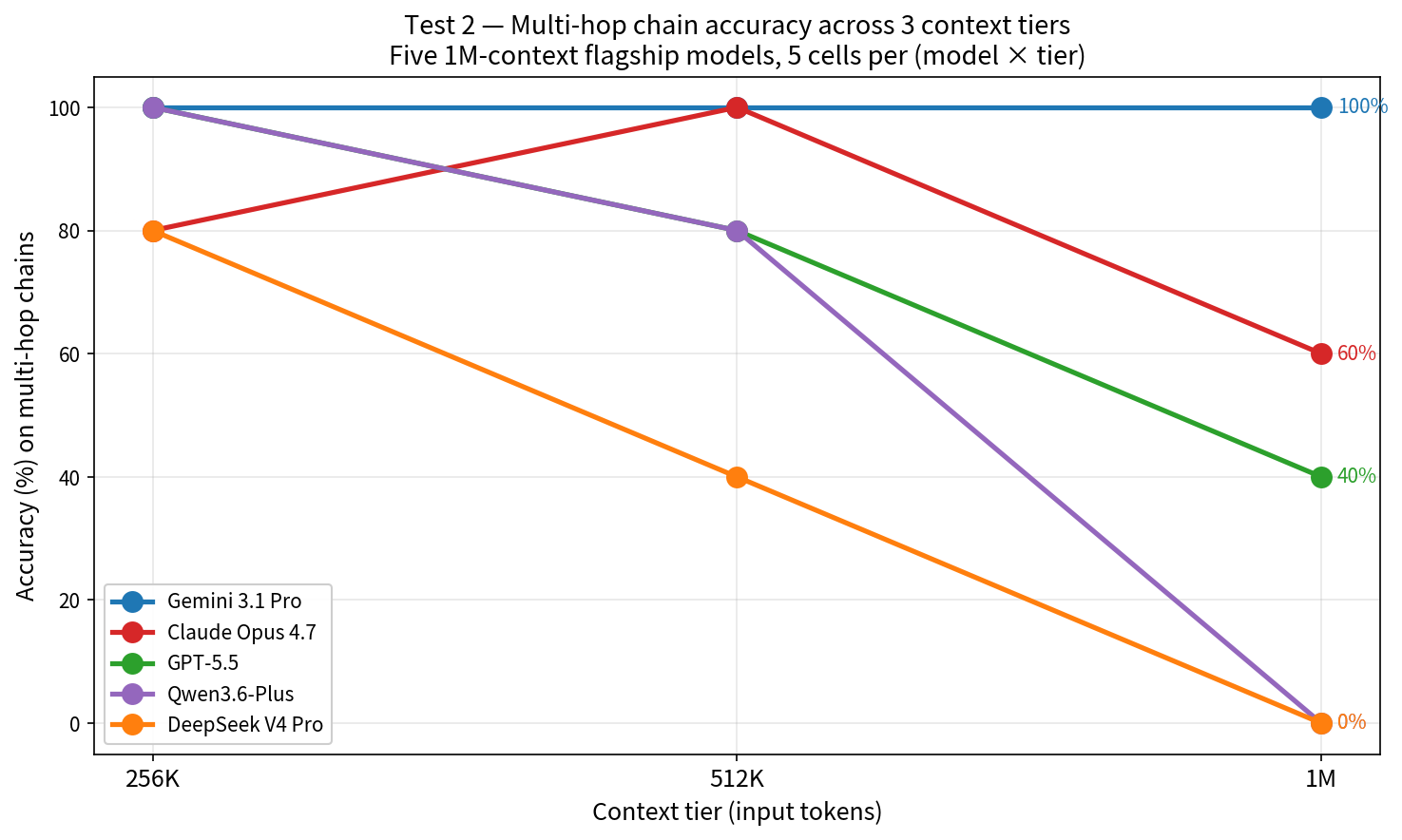}
\caption{Per-model accuracy across the three context tiers.}
\end{figure}

\subsection{Three Decay Signatures}

Inspection of the per-tier curves (Figure~3) reveals three qualitatively distinct decay shapes:

\begin{itemize}[leftmargin=*]
\item \textbf{Stable regime} (Gemini 3.1 Pro, Claude Opus 4.7). Accuracy remains at $\geq$4/5 through the 256K and 512K tiers, with a moderate drop at 1M only. Gemini Pro achieves a perfect 5/5 across all three tiers. Claude shows some variability, dropping to 3/5 at 1M but with substantial run-to-run sampling noise .

\item \textbf{Late-cliff regime} (GPT-5.5, Qwen3.6-plus). Both models maintain near-perfect accuracy through 512K and then collapse abruptly at 1M. GPT-5.5 transitions from 4/5 to 2/5 across the single 512K\,$\to$\,1M tier jump; Qwen3.6-plus from 4/5 to 0/5. The transition is far steeper than predicted by smooth interpolation between adjacent tiers.

\item \textbf{Smooth-decline regime} (DeepSeek V4 Pro). Accuracy decreases steadily: 4/5\,$\to$\,2/5\,$\to$\,0/5 across 256K\,$\to$\,512K\,$\to$\,1M. No sharp transition is apparent; instead the model loses capability gradually across the range.
\end{itemize}

\subsection{Per-Chain Analysis}

The five chains query different attribute types, and performance varies substantially across them (Figure~4 and Table~7). Chain~C3 (hometown) is the hardest: only Gemini 3.1 Pro solves it at all three tiers, while every other model fails at 1M. Chain~C5 (courtesy name, 字) is similarly difficult, with DeepSeek V4 Pro scoring 0/3 even when including the 256K tier. By contrast, Chain~C1 (age at death) and Chain~C4 (book authored) are the easiest, with three models achieving 3/3.

\begin{table}[h]
\centering
\small
\begin{tabular}{@{}lrrrrr@{}}
\toprule
\textbf{Model} & \textbf{C1 (age)} & \textbf{C2 (year)} & \textbf{C3 (place)} & \textbf{C4 (book)} & \textbf{C5 (字)} \\
\midrule
Gemini 3.1 Pro     & 3/3 & 3/3 & 3/3 & 3/3 & 3/3 \\
Claude Opus 4.7    & 3/3 & 2/3 & 1/3 & 3/3 & 3/3 \\
GPT-5.5            & 3/3 & 2/3 & 1/3 & 3/3 & 2/3 \\
Qwen3.6-Plus       & 2/3 & 2/3 & 1/3 & 2/3 & 2/3 \\
DeepSeek V4 Pro    & 2/3 & 1/3 & 1/3 & 2/3 & 0/3 \\
\bottomrule
\end{tabular}
\caption{Per-chain accuracy pooled across the three context tiers (3 cells per entry = one cell at each of 256K, 512K, 1M).}
\end{table}

\begin{figure}[h]
\centering
\includegraphics[width=0.95\textwidth]{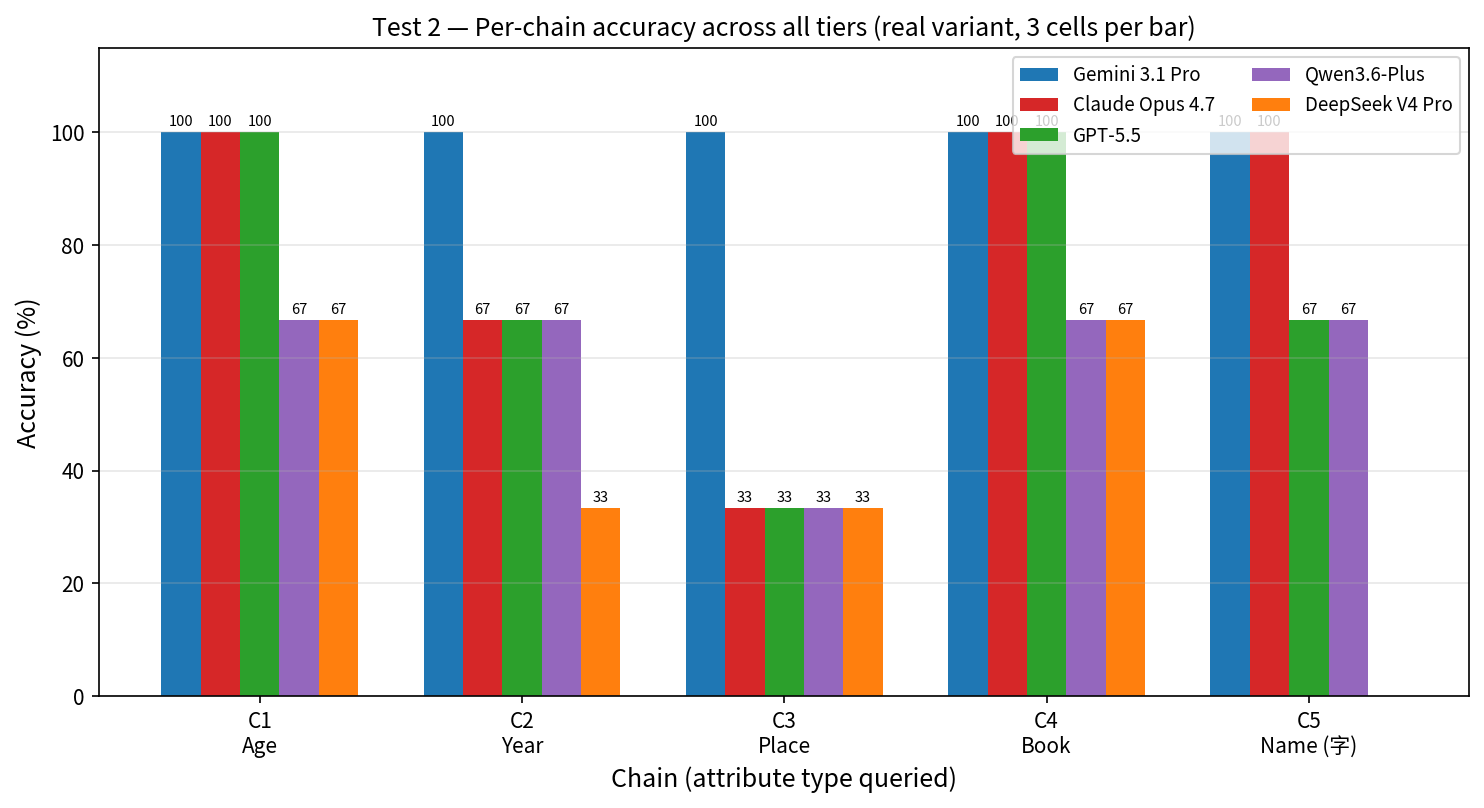}
\caption{Per-chain accuracy across all tiers. Chain~C3 (hometown) is the hardest chain; Chain~C5 (courtesy name) is notably difficult for DeepSeek V4 Pro. Only Gemini 3.1 Pro achieves 100\% on every chain.}
\end{figure}

This variation suggests that not all multi-hop queries are equally hard: chains whose target attribute requires disambiguating between structurally similar classical-Chinese terms (e.g., distinguishing a person's name from their courtesy name in Chain~C5, or identifying a friend's hometown rather than the teacher's in Chain~C3) are more prone to failure than chains whose target is a distinctive token (e.g., a book title in Chain~C4).

\subsection{Failure Mode Taxonomy}

We catalog five recurrent failure modes observed at the 1M tier:

\begin{enumerate}[label=\arabic*.,leftmargin=*]
\item \textbf{Subject substitution hallucination.} The model conflates a fictional chain subject with a similarly-named real Song-era figure from its training data. \emph{Example:} GPT-5.5 on Chain~C5~--- ``黃師復 應是指 管師復'' (then proceeds to describe Guan Shifu's biography). Most prevalent in GPT-5.5 and Qwen3.6-plus.

\item \textbf{Wrong-attribute selection.} The model traverses the chain but reports an attribute of the wrong entity. \emph{Example:} DeepSeek V4 Pro consistently reports the \emph{teacher's} zì on Chain~C5, rather than the friend's zì (the chain's intended target), across all four Chain~C5 cells.

\item \textbf{Truncated chain.} The model identifies the first hop and reports an answer at that level, never reaching the chain's intended endpoint. \emph{Example:} Qwen3.6-plus on Chain~C3 returns the teacher's hometown rather than the friend's hometown.

\item \textbf{Confident attribute fabrication.} The model produces a plausible-sounding answer that has no basis in the haystack. \emph{Example:} GPT-5.5 on Chain~C4 invents the book title ``詩筌'' (a real Song-era poetic treatise unrelated to the question).

\item \textbf{Refusal.} The model declines to answer, sometimes citing absence of information in the haystack, sometimes invoking compliance constraints. \emph{Example:} DeepSeek V4 Pro on Chain~C2 at 512K returns ``對不起，我不能提供和評價這類信息'' (``Sorry, I cannot provide or evaluate this type of information''~--- a generic compliance refusal). The same response wording recurs across runs.
\end{enumerate}

\section{Discussion}

\subsection{The 512K $\to$ 1M Cliff}

The single most striking feature of Test 2's data is the sharp transition observed for GPT-5.5 and Qwen3.6-plus between the 512K and 1M tiers. Both models advertise context windows of approximately 1M tokens (922K input for GPT-5.5; 983K usable for Qwen3.6-plus). Their 512K performance is strong~--- 4/5 each~--- yet at 1M they collapse to 2/5 and 0/5. This pattern echoes RULER's finding that claimed context size substantially overstates effective capability (Hsieh et al., 2024), but extends it to the 1M-token frontier and to multi-hop reasoning on a real-world corpus rather than synthetic variable-tracking tasks.

By contrast, Gemini 3.1 Pro and Claude Opus 4.7 maintain $\geq$3/5 accuracy at 1M, suggesting that their long-context attention scales more uniformly across the advertised window. BABILong (Kuratov et al., 2024) reported that popular LLMs utilise only 10--20\% of their context; our stable-regime models appear to be exceptions to this pattern, at least for the retrieval-plus-reasoning tasks tested here.

\subsection{Tokenization and Effective Context}

The factor-of-1.35 spread in tokens-per-character across model tokenizers (DeepSeek 0.901 to Claude 1.222) implies that for an identical character-level corpus, Claude's 1M-token budget admits 23\% less \emph{content} than DeepSeek's. This tokenization asymmetry is a confound for any cross-model comparison that does not control for it. Our per-model haystack construction normalizes on token count (each model is tested at $\sim$99\% of its own context budget), but this implies the \emph{content} differs across models. The alternative (matching content and accepting different token budgets) merely shifts the methodological compromise. We adopt token normalization as the more interpretable choice, given that long-context attention is a token-level phenomenon.

\subsection{Separating Retrieval from Memorised Training Data}

Test 1's altered-variant design exposes two distinct behaviors. Models that successfully retrieve from the provided text (Gemini Pro, Claude, GPT-5.5) achieve identical scores on real and altered variants. Models that fall back to memorised training data (Qwen3.6-plus, partially DeepSeek V4 Pro) show asymmetric scores: high on real, low on altered. This asymmetry is the cleanest empirical separator of ``reading the text'' from ``reciting from memory'' in the absence of more elaborate behavioral probes. Future evaluations of long-context models on culturally significant corpora should use paired real/altered designs by default.

\subsection{Implications for Practical Deployment}

The decay signatures observed in Test 2 have direct implications for the RAG-versus-long-context debate (Li et al., 2025a; Li et al., 2025b). Where prior work concluded that ``neither RAG nor long-context LLMs are a silver bullet'' (LaRA), our results add a finer-grained picture: the answer depends not only on the task but on the specific \emph{model} and the specific \emph{context tier}. A RAG system targeting 1M-context inference must factor in not only the advertised context window but the \emph{shape} of accuracy decay within it. For Gemini 3.1 Pro and Claude Opus 4.7, performance degrades gracefully and partial reliance on long context is reasonable. For GPT-5.5 and Qwen3.6-plus, performance is acceptable through approximately 512K input tokens but drops sharply beyond that point; deployments using these models should target context utilization at or below the 512K range. For DeepSeek V4 Pro, smooth decline through the entire range argues for shorter effective contexts when accuracy is critical.

For multi-hop questions specifically~--- the kind of query Test 2 measures~--- the same decay signatures predict where an explicit graph-based retrieval stage would help. As discussed in §\ref{sec:kg}, the chain task is structurally identical to a GraphRAG traversal query: three triples, one path. Stable-regime models (Gemini Pro, Claude) already resolve the implicit graph from the haystack with high reliability, so a preprocessing stage that materialises an explicit knowledge graph would offer little marginal benefit at 1M. Late-cliff models (GPT-5.5, Qwen3.6-plus) would benefit substantially: their 512K performance is near-perfect, suggesting the limitation is not graph-extraction quality but long-context attention to extracted relations. A GraphRAG architecture in front of these models~--- which compresses the haystack into a compact relation set before LLM inference~--- would likely move their 1M-equivalent task performance into the stable regime. Smooth-decline models (DeepSeek V4 Pro) would benefit at every tier, because the implicit construction stage is already fragile at 256K. The Test 2 results thus operationalise the \emph{when-to-use-GraphRAG} decision as a function of model identity and context length. For historians querying million-character corpora of the kind tested here, the practical implication is that a ``dump the full text into the prompt'' strategy works reliably for simple lookups on the strongest models, but multi-hop prosopographical queries~--- tracing lineages, identifying indirect relationships between scholars~--- require either a model in the stable regime or an explicit knowledge-graph layer in front of a weaker one.

\subsection{Cost Considerations: Prompt Caching}

Since all five models are accessed via cloud APIs, the cost of sending a million-token corpus with every query is a practical concern. All five providers now offer some form of \emph{prompt caching}: the corpus is ``read'' and stored on the first query, and subsequent queries against the same corpus pay a substantially reduced input rate. Two providers (GPT-5.5 and DeepSeek V4 Pro) enable this automatically with no code changes; two (Gemini 3.1 Pro and Claude Opus 4.7) require the developer to mark cacheable content explicitly in the API call; one (Qwen3.6-plus) documents the feature but with less clarity on the exact mechanism. The discounts are significant~--- ranging from 50\% (GPT-5.5) to over 90\% (Claude, DeepSeek) off the standard input rate for cached portions.

For the ``dump the full corpus and ask questions'' workflow, this means the first query pays full price but each subsequent query against the same corpus costs a fraction. A historian posing twenty questions against the full text of the 宋元學案 would pay roughly the cost of two or three full-price queries rather than twenty. This makes the long-context approach economically viable for iterative research, even at 1M-token input scales~--- provided the researcher batches queries within the cache's time-to-live window (typically five minutes to twenty-four hours, depending on the provider).

\subsection{Scope and Limitations}

We emphasise that the present study is an exploratory pilot, not a benchmark. The sample sizes are small~--- 18 cells per model in Test 1 and 15 in Test 2~--- and each cell is queried only once. The three decay signatures we identify are qualitative patterns that emerge clearly from the data, but the per-model accuracy figures should not be read as precise point estimates. A full benchmark would require substantially more chains, more depth positions, multiple independent runs per cell, and ideally a held-out validation set of chains not seen during design. We present these results as a methodological proof-of-concept and a set of directional findings that motivate such a benchmark, rather than as definitive model rankings.

With that framing, we note the following specific limitations:

\begin{enumerate}[label=\arabic*.,leftmargin=*]
\item \textbf{Single-domain corpus.} Our findings are specific to classical Chinese; we make no claim that the same decay signatures apply to English, modern Chinese, or other domains.
\item \textbf{Non-uniform determinism across models.} Three of the five models (Gemini 3.1 Pro, Qwen3.6-plus, DeepSeek V4 Pro) support temperature\,=\,0 and produce deterministic outputs. The remaining two (Claude Opus 4.7, GPT-5.5) enforce temperature\,=\,1 when reasoning mode is enabled, meaning their results are inherently non-deterministic single samples. Future work should run multiple trials on these models and report variance; extending this to all models at varying reasoning-effort levels would further clarify whether the observed decay signatures are stable properties of the models or artefacts of a particular sampling regime.
\item \textbf{No human evaluation of replies.} Substring matching captures whether the expected fact appears in the model's response, but does not penalize hallucinated supporting reasoning around a correct answer. Manual inspection of replies suggests this is rare for the strongest models but does occur for weaker ones; a human-graded subset would tighten the analysis.
\item \textbf{Limited chain depth.} Test 2 uses three-hop chains exclusively. Two-hop chains may distinguish models that fail Test 2 entirely; four- or five-hop chains may distinguish among the stable-regime models.
\item \textbf{No tested intermediate tier between 512K and 1M.} The sharpness of the GPT-5.5 / Qwen3.6-plus transition is established between 512K and 1M, but its precise breakpoint is undetermined.
\end{enumerate}

\section{Conclusion}

We measured single-needle retrieval and three-hop chain reasoning across five 1M-context language models on a classical Chinese corpus. Single-needle retrieval at 1M tokens is solved for the strongest current models (Gemini 3.1 Pro, Claude Opus 4.7, GPT-5.5, all 18/18). Multi-hop chain reasoning, which requires traversing intermediate references rather than locating a single fact, reveals three distinct decay signatures across the 256K\,$\to$\,512K\,$\to$\,1M tier range. Two models (GPT-5.5, Qwen3.6-plus) exhibit a sharp 512K\,$\to$\,1M cliff, suggesting that nominal context-window length substantially overstates their usable long-context capability. Two models (Gemini 3.1 Pro, Claude Opus 4.7) maintain stable performance through 1M with manageable degradation. One model (DeepSeek V4 Pro) shows smooth decline across the entire range.

Prior benchmarks have established that single-needle retrieval is largely solved (Kamradt, 2023; Hsieh et al., 2024), that LLMs utilise only a fraction of their advertised context for reasoning tasks (Kuratov et al., 2024), and that the choice between RAG and long context depends on task and model (Li et al., 2025a; Li et al., 2025b). Our results extend these findings in three ways. First, we show that the gap between advertised and effective context persists at the 1M-token frontier, not just at the 32K--128K range tested previously. Second, we demonstrate that the \emph{shape} of decay --- cliff, smooth, or stable --- varies across models, providing actionable guidance that binary ``RAG or long context'' comparisons do not. Third, we show that these patterns hold on a non-English, real-world historical corpus where parametric leakage, tokenization asymmetry, and domain-specific ambiguity introduce challenges absent from synthetic English benchmarks.

For digital-humanities researchers working with classical Chinese corpora, the practical takeaway is twofold: simple factual lookups can be reliably handled by long-context inference on the strongest models, but multi-hop prosopographical queries benefit from either choosing a stable-regime model or adding a GraphRAG layer. Our paired real/altered needle design offers a methodology for separating genuine in-context retrieval from reliance on memorised training data in domains where such overlap is unavoidable.

\section*{References}
\begingroup
\setlength{\parindent}{0pt}
\setlength{\parskip}{0.5em}
\everypar={\hangindent=2em \hangafter=1}

Cao, J., Liu, Y., Shi, Y., Ding, K., \& Jin, L. (2024). WenMind: A Comprehensive Benchmark for Evaluating Large Language Models in Chinese Classical Literature and Language Arts. \emph{Advances in Neural Information Processing Systems 37 (NeurIPS 2024) Datasets and Benchmarks Track}.

Edge, D., Trinh, H., Cheng, N., Bradley, J., Chao, A., Mody, A., Truitt, S., Metropolitansky, D., Ness, R. O., \& Larson, J. (2024). From Local to Global: A GraphRAG Approach to Query-Focused Summarization. \emph{arXiv preprint} arXiv:2404.16130.

Hsieh, C.-P., Sun, S., Kriman, S., Acharya, S., Rekesh, D., Jia, F., Zhang, Y., \& Ginsburg, B. (2024). RULER: What's the Real Context Size of Your Long-Context Language Models? \emph{Conference on Language Modeling (COLM 2024)}. arXiv:2404.06654.

Kamradt, G. (2023). Needle In A Haystack --- Pressure Testing LLMs. GitHub repository, \url{https://github.com/gkamradt/LLMTest_NeedleInAHaystack}.

Kuratov, Y., Bulatov, A., Anokhin, P., Rodkin, I., Sorokin, D., Sorokin, A., \& Burtsev, M. (2024). BABILong: Testing the Limits of LLMs with Long Context Reasoning-in-a-Haystack. \emph{Advances in Neural Information Processing Systems 37 (NeurIPS 2024) Datasets and Benchmarks Track}. arXiv:2406.10149.

Li, K., Zhang, L., Jiang, Y., Xie, P., Huang, F., Wang, S., \& Cheng, M. (2025b). LaRA: Benchmarking Retrieval-Augmented Generation and Long-Context LLMs --- No Silver Bullet for LC or RAG Routing. \emph{Proceedings of the 42nd International Conference on Machine Learning (ICML 2025)}. arXiv:2502.09977.

Li, M., Zhang, S., Zhang, T., Duan, H., Liu, Y., \& Chen, K. (2024). NeedleBench: Evaluating LLM Retrieval and Reasoning Across Varying Information Densities. \emph{Transactions on Machine Learning Research (TMLR), 09/2025}. arXiv:2407.11963.

Li, X., Cao, Y., Ma, Y., \& Sun, A. (2025a). Long Context vs.\ RAG for LLMs: An Evaluation and Revisits. \emph{arXiv preprint} arXiv:2501.01880.

Yao, X., Wang, M., Chen, B., \& Zhao, X. (2025). WenyanGPT: A Large Language Model for Classical Chinese Tasks. \emph{arXiv preprint} arXiv:2504.20609.

Yu, Y., Zhang, Q.-W., Qiao, L., Yin, D., Li, F., Wang, J., Chen, Z., Zheng, S., Liang, X., \& Sun, X. (2025). Sequential-NIAH: A Needle-In-A-Haystack Benchmark for Extracting Sequential Needles from Long Contexts. \emph{arXiv preprint} arXiv:2504.04713.

Zhou, B., Chen, Q., Wang, T., Zhong, X., \& Zhang, Y. (2023). WYWEB: A NLP Evaluation Benchmark For Classical Chinese. \emph{Findings of the Association for Computational Linguistics: ACL 2023}. arXiv:2305.14150.

\endgroup

\end{document}